# Second-Order Component Analysis for Fault Detection

Jingchao Peng[1], Haitao Zhao[2], and Zhengwei Hu[3]
East China University of Science and Technology,
Automation Department, School of Information Science and Engineering
{[1]pjc, [3]zwh}@mail.ecust.edu.cn, [2]htz@ecust.edu.cn

***Abstract*—Process monitoring based on neural networks is getting more and more attention. Compared with classical neural networks, high-order neural networks have natural advantages in dealing with *heteroscedastic* data. However, high-order neural networks might bring the risk of overfitting and learning both the key information from original data and noises or anomalies. Orthogonal constraints can greatly reduce correlations between extracted features, thereby reducing the overfitting risk. This paper proposes a novel fault detection method called second-order component analysis (SCA). SCA rules out the heteroscedasticity of process data by optimizing a second-order autoencoder with orthogonal constraints. In order to deal with this constrained optimization problem, a geometric conjugate gradient algorithm is adopted in this paper, which performs geometric optimization on the combination of Stiefel manifold and Euclidean manifold. Extensive experiments on the Tennessee-Eastman benchmark process show that SCA outperforms PCA, KPCA, and autoencoder in missed detection rate (MDR) and false alarm rate (FAR).**

***Index Terms*—Fault detection, process monitoring, high-order neural network, orthogonal constraint, Riemannian manifold**

## I. Introduction

CHEMICAL processes bring irreparable damages in the event of a fault or accidental shutdown. Fault detection methods are widely adopted to deal with these damages, especially in large-scale, high-complexity, multi-variable modern industrial processes to reduce downtime, minimize manufacturing costs, and improve equipment safety [1]-[3].

Due to the widespread applications of modern digital instruments and distributed control systems, more and more data are available [4]. Utilizing these data for fault detection can effectively prevent accidents and faults, which is of great significance to the safety of chemical processes. However, in the actual production processes, the data obtained often contains a large number of correlated variables [5], which leads to "curse of dimensionality" and "data rich but information poor" problems [6]. In order to deal with these problems, fault detection usually designs a dimension reduction method to extract the key features and then perform fault detection in low-dimensional feature space [7]. Traditional feature extraction methods, such as principal component analysis (PCA) [6]-[9] and autoencoders [10]-[13], have been widely applied in chemical processes, biochemical processes, and semiconductor processes.

Principal component analysis (PCA) tries to find the significant features of the data through a linear transformation. One essential characteristic of PCA is that it has orthogonal constraints, reducing the correlations between extracted features [8], [9]. PCA projects the raw data into two subspaces: a significant subspace that contains key features for the reconstruction of the training data and a residual subspace which contains noises or anomalies. However, the linear transformation obtained by PCA is difficult to tackle the nonlinear relationship between different process variables, and the PCA-based methods may lead to unreliable and ineffective fault detection in monitoring nonlinear processes [14], [15].

In order to monitor nonlinear processes, autoencoders are widely used [3], [10]-[13]. The model of a classical autoencoder is divided into two parts: an encoder and a decoder. The encoder attempts to learn the features of the original data, and the decoder implements data reconstruction by minimizing reconstruction errors. After the learning process, only the trained encoder is utilized to extract features for further fault detection. In the past decade, with intense research on deep learning, autoencoder-based methods have been widely used [3], [10] in fault detection. In order to better reconstruct high-complex chemical processes, automatic encoders tend to be larger and deeper. A larger and deeper network represents better results. However, on the other hand, it may make the model more complicated [16], which can reduce the generalization ability of the model and increase the risk of overfitting.

Inspired by the ultrastructure of biological nervous systems, researchers proposed high-order neural network models. High-order neural networks do not simply increase the number of neurons but use summers and integrators to construct high-order neurons, which can greatly increase the nonlinear mapping capabilities of the network [17].

However, due to the lack of orthogonal constraints, autoencoder-based methods are prone to learn both the key information from original data and noises or anomalies [18]. These methods are easier to overfit the training data since they tend to capture all information than to reduce the correlation [19].

Based on the above analysis, this paper proposes a novel fault detection method called second-order component analysis (SCA). SCA has an encoder-decoder structure. The encoder of SCA is a feedforward neural network to encode the original data into the feature space, which improves the expression ability of the shallow autoencoder by using a second-order neural network. The decoder is a projection with orthogonal constraints to decode the features. The orthogonal constraints reduce the risk of overfitting the training data.

Section II introduces related technologies, including PCA, autoencoder, and high-order neural networks. Section III proposes our newly designed feature extraction method, SCA, and demonstrates its effectiveness through a motivation example. Section IV introduces the fault detection method based on SCA and elaborates the offline training and online monitoring procedures. In Section V, the Tennessee-Eastman (TE) process is utilized to show the superiority and feasibility of SCA. Section VI presents the conclusions.

## II. Related Work

### A. Principal Component Analysis

Principal component analysis is one of the most widely used fault detection methods, which transforms correlated raw data into significant features by minimizing the reconstruction error under orthogonal constraints. The optimization problem of principal component analysis is:

$$W_{PCA} = \underset{W}{\operatorname{argmin}} \|X - WW^TX\|_F^2 \quad s.t.\ W^TW = I_{p\times p} \tag{1}$$

Where $X = [\boldsymbol{x}^{(1)}, \boldsymbol{x}^{(2)}, \cdots, \boldsymbol{x}^{(m)}] \in \mathbb{R}^{n\times \mathrm{m}}$ is the training data, $W_{PCA} = [\boldsymbol{w}^{(1)}, \cdots, \boldsymbol{w}^{(p)}] \in \mathbb{R}^{n\times p}$ is a column-orthogonal matrix. $\|\cdot\|_F$ denotes the Frobenius norm. The solution to (1) is an orthogonal transformation formed by $p (p < n)$ eigenvectors corresponding to the first $p$ eigenvalues of the sample covariance matrix $\Sigma_x$. In PCA, $G = {W_{PCA}}^T X$ is often called the score matrix corresponding to the transformation $W_{PCA}$ ($W_{PCA}$ is also called the loading matrix).

Due to the transformation $W_{PCA}$, PCA can transform the original space into two orthogonal subspaces. One is the significant subspace, which contains the key features for the reconstruction of original data. The other is the residual subspace, which includes noises or anomalies in training data. Orthogonal transformation minimizes the correlation in the original data. Due to this orthogonal characteristic, PCA and its extensions are widely used for process monitoring [6], [20].

### B. Autoencoder

Autoencoder is a neural network for unsupervised feature extraction. AE can be divided into two parts: the encoder and the decoder. It tries to learn the encoding of the original data and realize feature extraction by minimizing the reconstruction error. Recently, due to the booming of deep learning, autoencoder-based fault detection methods have been widely studied [12], [13].

The simplest autoencoder (called AE in this paper) consists of an input layer of $n$ nodes, a hidden layer of $p$ nodes, and an output layer of $n$ nodes, illustrated in Fig. 1. The purpose of this structure is to reconstruct the input. It is easy to find that AE belongs to unsupervised learning.

Autoencoder includes an encoder and a decoder, which can be defined as a mapping $\alpha$ and a mapping $\beta$ respectively, i.e.:

$$\alpha: \mathbb{R}^{n\times \mathrm{m}} \to \mathcal{F}$$
$$\beta: \mathcal{F} \to \mathbb{R}^{n\times \mathrm{m}}$$
$$[\alpha, \beta] = \underset{\alpha,\beta}{\operatorname{argmin}} \sum_{i=1}^{n} \left\|\boldsymbol{x}_i - \beta\big(\alpha(\boldsymbol{x}_i)\big)\right\|^2 \tag{2}$$

The encoder receives the input $X$ and maps it to $G \in \mathcal{F}$:

$$G = \alpha(X) \triangleq \sigma(W^TX + \boldsymbol{b}) \tag{3}$$

This feature matrix $G$ is called feature codes, potential features or potential representations, $\sigma(\cdot)$ is an element activation function, such as a sigmoid function or a hyperbolic tangent function, $W$ is a parameter matrix, $\boldsymbol{b}$ is an offset vector.

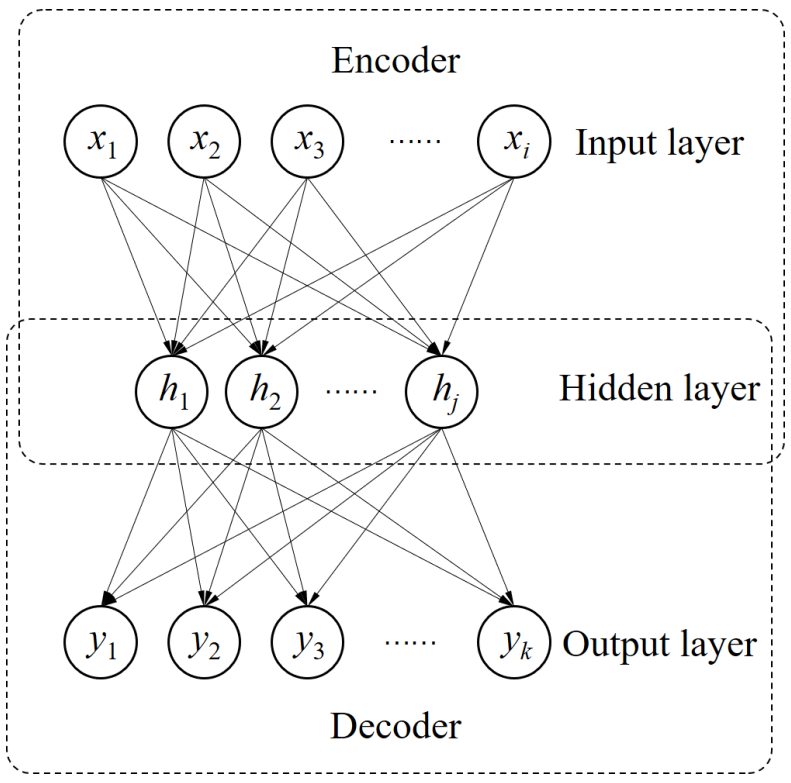

Fig. 1. Structure of autoencoder.

The decoder maps the feature codes to the reconstructed output $\tilde{X}$ of the same dimension as $X$:

$$\tilde{X} = \beta(G) \triangleq \tilde{\sigma}\big(\widetilde{W}G + \tilde{\boldsymbol{b}}\big) \tag{4}$$

The $\tilde{\sigma}$, $\widetilde{W}$, and $\tilde{\boldsymbol{b}}$ of the decoder may be different from the $\sigma$, $W$ and $\boldsymbol{b}$ of the encoder, depending on the applications of the autoencoder.

To learn the parameters $\big[W, \boldsymbol{b}; \widetilde{W}, \tilde{\boldsymbol{b}}\big]$, autoencoder is usually trained to minimize the reconstruction error:

$$\big[W, \boldsymbol{b}; \widetilde{W}, \tilde{\boldsymbol{b}}\big] = \underset{W,\boldsymbol{b},\widetilde{W},\tilde{\boldsymbol{b}}}{\operatorname{argmin}} \sum_{i=1}^{n} \left\|\boldsymbol{x}_i - \tilde{\sigma}\big(\widetilde{W}\sigma(W^TX + \boldsymbol{b}) + \tilde{\boldsymbol{b}}\big)\right\|^2 \tag{5}$$

After the training is completed, only the encoder is utilized to obtain the feature codes.

### C. High-Order Neural Networks

High-order neural networks (HONN) are feedforward neural networks whose input layer contains input terms and high-order expansion terms. The high-order expansion terms are the polynomial expansion of the input vector $\boldsymbol{x}^{(i)} = \big[x_1^{(i)}, x_2^{(i)}, \cdots, x_n^{(i)}\big]$, for example $\big(x_j^{(i)}x_k^{(i)}, x_j^{(i)}x_k^{(i)}x_l^{(i)}, \dots\big)$, where $(j, k, l)$ are integers between 1 and $n$, respectively. HONN networks with order $k$ $(k \geq 2)$ are called $k$th-order neural networks.

Chemical processes are usually very complicated, and the data often has characteristics such as strong nonlinearity and heteroscedasticity [21]. High-order neural networks can provide nonlinear mapping and have natural advantages in dealing with heteroscedastic data. The following toy example shows that the boundary of heteroscedastic data is determined by the square term as well as the primary terms.

**Toy Example:** Consider a simple encoder that includes an input layer of $n$ nodes and a hidden layer of 1 node to get the code (or feature) of the input $n$-dimensional input data. In this case, the encoder receives an input $\boldsymbol{x} \in \mathbb{R}^n$ and map it to $y \in \{0,1\}$: $y = \alpha(\boldsymbol{x}) = \sigma(W\boldsymbol{x} + \boldsymbol{b})$.

In this example, $y = 0$ means $\boldsymbol{x}$ is normal data, and $y = 1$ means $\boldsymbol{x}$ is abnormal data. Let the density function of the normal data be:

$$p(\boldsymbol{x}|y=0) = \mathcal{N}(\boldsymbol{x}|\boldsymbol{\mu}_1, \sigma_0) = \frac{1}{(2\pi)^{n/2}{\sigma_0}^n}\exp\left(-\frac{1}{2{\sigma_0}^2}(\boldsymbol{x}-\boldsymbol{\mu}_0)^T(\boldsymbol{x}-\boldsymbol{\mu}_0)\right) \tag{6}$$

Moreover, the density function of the abnormal data be:

$$p(\boldsymbol{x}|y=1)=\mathcal{N}(\boldsymbol{x}|\boldsymbol{\mu}_1,\sigma_1)$$
$$=\frac{1}{(2\pi)^{n/2}{\sigma_1}^n}\exp\left(-\frac{1}{2{\sigma_1}^2}(\boldsymbol{x}-\boldsymbol{\mu}_1)^T(\boldsymbol{x}-\boldsymbol{\mu}_1)\right) \quad (7)$$

where $\boldsymbol{\mu}_1 \neq \boldsymbol{\mu}_2$ and $\sigma_1 \neq \sigma_2$. For simplicity, assume the priors are equal, i.e., $P(y=0)=P(y=1)$. If Bayesian inference is adopted to obtain the posterior probability, we have

$$P(y=0|x)=\frac{p(\boldsymbol{x}|y=0)p(y=0)}{p(\boldsymbol{x}|y=0)p(y=0)+p(\boldsymbol{x}|y=1)p(y=1)}$$
$$=\frac{1}{(1+(\sigma_0/\sigma_1)^n\exp(a\boldsymbol{x}^T\boldsymbol{x}-b\boldsymbol{x}+c))} \quad (8)$$

where $a=1/2{\sigma_0}^2-1/2{\sigma_1}^2$, $b={\boldsymbol{\mu}_1}^T/{\sigma_1}^2-{\boldsymbol{\mu}_0}^T/{\sigma_0}^2$, and $c={\boldsymbol{\mu}_0}^T\boldsymbol{\mu}_0/2{\sigma_0}^2-{\boldsymbol{\mu}_1}^T\boldsymbol{\mu}_1/2{\sigma_1}^2$.

From (8), we can find that when $\sigma_0=\sigma_1$, i.e., both the input variables of the normal data and those of the abnormal data have the same variance, the posterior probability $P(0\,|\,x)=1/(1+\exp(-(\boldsymbol{w}^T\boldsymbol{x}+b)))$, where $w=1/\sigma^2(\boldsymbol{\mu}_0-\boldsymbol{\mu}_1)$, $b=-(\boldsymbol{\mu}_0^T\boldsymbol{\mu}_0)/(2\sigma_0^2)+(\boldsymbol{\mu}_1^T\,\boldsymbol{\mu}_1)/(2\sigma_1^2)$. Please note that in this case, $P(0|x)$ is a sigmoid function. In this case, the Bayesian inference can be considered as an encoder with the sigmoid activation function.

However, when $\sigma_0 \neq \sigma_1$, in (8), $(1/2{\sigma_1}^2-1/2{\sigma_2}^2)\boldsymbol{x}^T\boldsymbol{x} \neq 0$. It means that the encode is related not only to the input $\boldsymbol{x}$ but also to the squared term $\boldsymbol{x}^T\boldsymbol{x}$.

This simple example can bring us some enlightenment in the encoder stage of fault detection: the fault of the system may be determined not only by the input variables but also by the squared terms of the variables.

Although the larger $k$ is, the HONN networks have better nonlinear mapping ability, too large $k$ may cause overfitting and parameter explosion. Studies have shown that a second-order neural network is sufficient to form a global universal approximator [22]. This paper adopts a second-order neural network as the encoder for feature extraction.

## III. Second-Order Component Analysis

### A. Second-Order Component Analysis

This section proposes a novel method for fault detection called second-order component analysis (SCA). SCA uses a second-order autoencoder as an encoder to extract features from a collection of the zeroth-order item, first-order terms, and second-order terms $\mathfrak{X}=\left[\boldsymbol{x}^{(1)},\boldsymbol{x}^{(2)},\cdots,\boldsymbol{x}^{(m)}\right]^T \in \mathbb{R}^{(1+n+n^2)\times m}$, where $\boldsymbol{x}^{(i)}=\left[1,x_1^{(i)},\cdots,x_n^{(i)},x_1^{(i)}x_1^{(i)},\cdots,x_n^{(i)}x_n^{(i)}\right] \in \mathbb{R}^{(1+n+n^2)\times 1}$. $x_j^{(i)}x_k^{(i)}$ is the second-order polynomial expansion of the input vector $\boldsymbol{x}^{(i)}=\left[1,x_1^{(i)},\cdots,x_n^{(i)}\right] \in \mathbb{R}^{n\times 1}$. Fig. 2 shows a fully connected second-order network.

In SCA, orthogonal constraints are also adopted to handle the overfitting problem. Then the optimization problem of SCA is

$$[\alpha,\beta]=\underset{\alpha,\beta}{\operatorname{argmin}}\left\|\mathfrak{X}-\beta\big(\alpha(\mathfrak{X})\big)\right\|^2 \quad \text{s.t. } \widetilde{W}^T\widetilde{W}=I_p \quad (9)$$

where $\alpha(\cdot)$ denotes transformation that the encoder part receives input $\mathfrak{X}$ and map it to $G \in \mathbb{R}^{p\times m}$, $\beta(\cdot)$ represents the mapping of the decoder part:

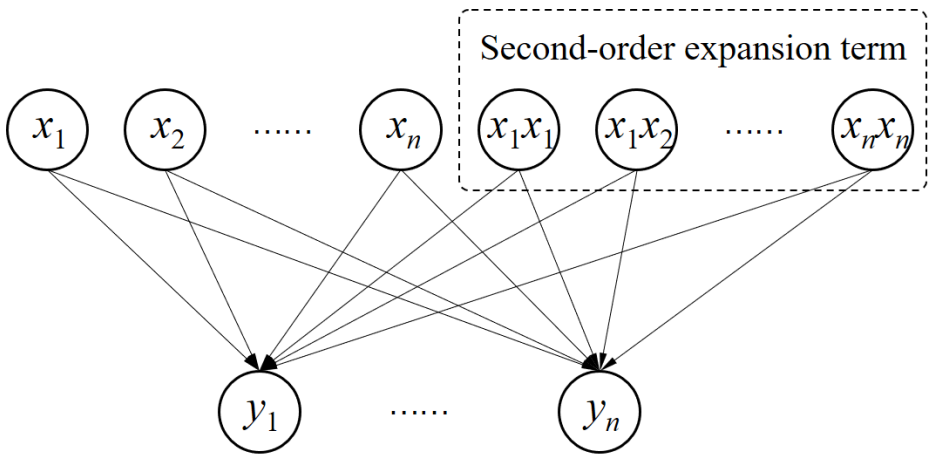

Fig. 2. Structure of second-order neural network.

$$G=\alpha(\mathfrak{X}) \triangleq \sigma(W^T\mathfrak{X}) \quad (10)$$
$$\widetilde{\mathfrak{X}}=\beta(G) \triangleq \tilde{\sigma}\big(\widetilde{W}G\big) \quad (11)$$

$\sigma(\cdot)$ and $\tilde{\sigma}(\cdot)$ are element activation function, such as the sigmoid function. Matrix $W$ and $\widetilde{W}$ are parametric matrix. The feature $G$ is called feature code, which is expected to be further analyzed in fault detection and detection.

Due to the orthogonal constraints $\widetilde{W}^T\widetilde{W}=I_{p\times p}$, the decoding from the underlying features is orthogonal, which can greatly reduce the correlation of different variables. However, orthogonal constraints also make (5) become a constrained optimization problem. In the following, we try to solve this optimization problem through the geometric conjugate gradient algorithm.

### B. SCA Optimization Problem Solution

The optimization problem of SCA can be rewritten as follows:

$$[W,\widetilde{W}]=\underset{W,\widetilde{W}}{\operatorname{argmin}}\left\|\mathfrak{X}-\tilde{\sigma}\left(\widetilde{W}\sigma(W^T\mathfrak{X})\right)\right\|_F^2 \quad \text{s.t. } \widetilde{W}^T\widetilde{W}=I_{p\times p} \quad (12)$$

Suppose $\widetilde{W}=\left[\widetilde{\boldsymbol{w}}^{(1)},\widetilde{\boldsymbol{w}}^{(2)},\cdots,\widetilde{\boldsymbol{w}}^{(p)}\right]$, where $\widetilde{\boldsymbol{w}}^{(i)} \in \mathbb{R}^{(1+n+n^2)\times 1}$, then the orthogonal constraints $\widetilde{W}^T\widetilde{W}=I_{p\times p}$ means:

$$\widetilde{\boldsymbol{w}}^{(i)^T}\widetilde{\boldsymbol{w}}^{(j)}=\begin{cases}1 & i=j\\ 0 & i\neq j\end{cases} \quad (i,j=1,2,\cdots,p) \quad (13)$$

In order to cope with orthogonal constraints, we employ the Stiefel manifold, which expresses all sets of orthogonal matrices. Let $\mathrm{St}(1+n+n^2,p)$ express $(1+n+n^2)\times p$ Stiefel manifold, i.e.:

$$\mathrm{St}(1+n+n^2,p)=\left\{\widetilde{W} \in \mathbb{R}^{(1+n+n^2)\times p}: \widetilde{W}^T\widetilde{W}=I_{p\times p}\right\} \quad (14)$$

It is easy to find that parameters to be optimized, $\widetilde{W}$, belong to the Stiefel manifold $\mathrm{St}(1+n+n^2,p)$. $W$ is an unconstrained parameter. Let Euclidean manifold $\mathrm{E}(1+n+n^2,p)$ be unconstrained Euclidean space of $(1+n+n^2)\times p$ matrices. Let the cost function by manifolds as:

$$f:\mathrm{St}(1+n+n^2,p)\times \mathrm{E}(1+n+n^2,p)\to\mathbb{R},$$
$$f\big(W,\widetilde{W}\big):=\left\|\mathfrak{X}-\tilde{\sigma}\left(\widetilde{W}\sigma(W^TZ)\right)\right\|_F^2 \quad (15)$$

The solution to the optimization problem in (12) is equivalent to minimize the cost function $f$ on the product manifold $\mathcal{M}=\mathrm{St}(1+n+n^2,p)\times \mathrm{E}(1+n+n^2,p)$.

In the following, we first propose a geometric conjugate gradient method on the manifold $\mathcal{M}$, cf. Algorithm 1. The diagram of the geometric conjugate gradient algorithm on the product manifold is shown in Fig. 3. Then we introduced the key techniques used in Algorithm 1, including retraction transformation and vector transport.

**Algorithm 1**: The geometric conjugate gradient algorithm on the product manifold.

---

**Input:** Initial iterate $\widetilde{W}_{(0)} \in \mathrm{St}(1+n+n^2,p)$ , $W_{(0)} \in \mathrm{E}(1+n+n^2,p)$, cost function $f$ as specified in (15)

**Output:** Accumulation point $(\widetilde{W}^*, W^*) \in \mathrm{St}(1+n+n^2,p) \times \mathrm{E}(1+n+n^2,p)$

**Step 1**: Set $k=1$, $\mathrm{M}_{(1)} = (\widetilde{W}_{(0)}, W_{(0)})$;

**Step 2**: Compute $G_{(1)} = H_{(1)} = -\operatorname{grad} f(\mathrm{M}_{(1)})$;

**Step 3**: set $k = k+1$;

**Step 4**: Update $M_{(k)} \leftarrow R_{\mathrm{M}_{(k-1)}}(t^* H_{(k-1)})$, where

$$t^* = \underset{t\in\mathbb{R}}{\operatorname{argmin}}\, R_{\mathrm{M}_{(k-1)}}(tH_{(k-1)})$$

**Step 5**: Update $G_{(k)} = -\operatorname{grad} f(M_{(k)})$;

**Step 6**: Update $H_{(k)} = G_{(k)} + \gamma \mathcal{T}_{M_{(k)}}(H_{(k-1)})$;

**Step 7**: If $\|M_{(k)} - M_{(k-1)}\|$ is small enough, stop. Otherwise, go to Step 3;

---

In Step 2 and Step 5, $\operatorname{grad} f(M_{(k)})$ means the Riemannian gradient on the point $M_{(k)}$:

$$\operatorname{grad} f(M_{(k)}) = (\operatorname{grad} f(\widetilde{W}_{(k)}), \operatorname{grad} f(W_{(k)})) \quad (16)$$

which refers to the projection of Euclidean gradient on the tangent space $T_{M_{(k)}}\mathcal{M}$. the tangent space of Euclidean manifold is itself. So, in Euclidean manifold $\mathrm{E}(1+n+n^2,p)$, the Riemannian gradient $\operatorname{grad} f(W_{(k)})$ is equal to the gradient $\nabla f(W_{(k)})$ on Euclidean space $\mathbb{R}^{(1+n+n^2)\times p}$, i.e.:

$$\operatorname{grad} f(W_{(k)}) = \nabla f(W_{(k)}) = -2\left(Z - \tilde{\sigma}\left(\widetilde{W}_{(k)}\sigma(W_{(k)}^T\mathfrak{X})\right)\right)\widetilde{W}_{(k)}^T\mathfrak{X} \quad (17)$$

While in Stiefel manifold $\mathrm{St}(1+n+n^2,p)$, the tangent space of Stiefel manifold is:

$$T_{\widetilde{W}}\mathrm{St}(1+n+n^2,p) = \{Z \in \mathbb{R}^{(1+n+n^2)\times p} : \widetilde{W}^T Z + Z^T \widetilde{W} = 0\} \quad (18)$$

And the Riemannian gradient $\operatorname{grad} f(\widetilde{W}_{(k)})$ is different from the gradient $\nabla f(\widetilde{W}_{(k)})$ on Euclidean space $\mathbb{R}^{(1+n+n^2)\times p}$, it can be calculated by:

$$\operatorname{grad} f(\widetilde{W}_{(k)}) = \nabla f(\widetilde{W}_{(k)}) - \frac{1}{2}\widetilde{W}_{(k)}\widetilde{W}_{(k)}^T \nabla f(\widetilde{W}_{(k)}) - \frac{1}{2}\widetilde{W}_{(k)} \nabla f(\widetilde{W}_{(k)})^T \widetilde{W}_{(k)} \quad (18)$$

where,

$$\nabla f(\widetilde{W}_{(k)}) = -2\left(Z - \tilde{\sigma}\left(\widetilde{W}_{(k)}\sigma(W_{(k)}{}^T Z)\right)\right)\left(\sigma(W_{(k)}{}^T Z)\right)^T \quad (19)$$

In Step 4, $t^*$ means search step, which can be obtained by one-dimensional search in the direction $\eta$:

$$t^* = \underset{t}{\operatorname{argmin}}\, f(t\eta) \quad (20)$$

The transformation $R.(\cdot)$ in Step 4 is a retraction, which is a smooth mapping from the tangent space $T_{M_{(k)}}\mathcal{M}$ to the manifold $\mathcal{M}$. In Euclidean manifold $\mathrm{E}(1+n+n^2,p)$, retraction transformation is an identity function $R.(\cdot) = \mathrm{id}.$ , i.e., $R_{W_{(k)}}(tW_{(k)}) = tW_{(k)}$. In Stiefel manifold $\mathrm{St}(1+n+n^2,p)$, retraction transformation makes every iteration point satisfies the orthogonal constraint, which can be calculated by:

$$R_{\widetilde{W}_{(k-1)}}(tH_{(k)}) = (\widetilde{W}_{(k-1)} + tH_{(k)})(I_{p\times p} + t^2 H_{(k)}{}^T H_{(k)})^{-\frac{1}{2}} \quad (21)$$
$$R_{\widetilde{W}_{(k-1)}}(tH_{(k)}) \in \mathrm{St}(1+n+n^2,p)$$

Theorem 1 proves that after the retraction transformation, the iteration points of each step satisfy the orthogonal constraints.

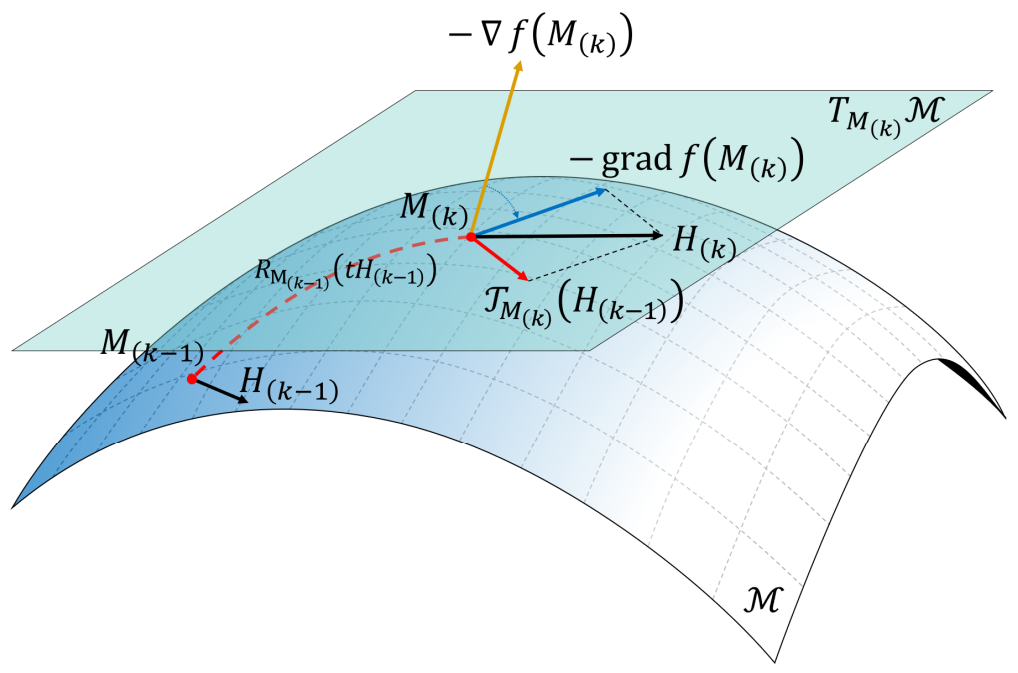

Fig. 3. Diagram of the geometric conjugate gradient algorithm on the product manifold.

**Theorem 1:** Let retraction transform be $R_W(\Xi) = (W+\Xi)(I_{p\times p} + \Xi^T\Xi)^\wedge(-1/2)$, The points $\Xi$ after retraction transform on the tangent space satisfies $R_W(\Xi)^T R_W(\Xi) = I_{p\times p}$.

**Proof.** The parameter after retraction $R_W(\Xi)$ satisfies:

$$R_W(\Xi)^T R_W(\Xi)$$
$$= \left((W+\Xi)(I_{p\times p} - \Xi^T\Xi)^{-\frac{1}{2}}\right)^T \left((W+\Xi)(I_{p\times p} + \Xi^T\Xi)^{-\frac{1}{2}}\right)$$
$$= (I_{p\times p} + \Xi^T\Xi)^{-\frac{1}{2}}(W+\Xi)^T(W+\Xi)(I_{p\times p} + \Xi^T\Xi)^{-\frac{1}{2}}$$
$$= (I_{p\times p} + \Xi^T\Xi)^{-\frac{1}{2}}((W+\Xi)^T(W+\Xi))(I_{p\times p} + \Xi^T\Xi)^{-\frac{1}{2}} \quad (22)$$

For $W \in \mathrm{St}(1+n+n^2,p)$, $\Xi$ satisfies:

$$W^T W = I_{p\times p} \quad (23)$$

For $\Xi$ is a point on the tangent space $T_W\mathrm{St}(1+n+n^2,p)$, $\Xi$ satisfies:

$$W^T\Xi + \Xi^T W = 0 \quad (24)$$

So

$$R_W(\Xi)^T R_W(\Xi) = (I_{p\times p} + \Xi^T\Xi)^{-\frac{1}{2}}(I_{p\times p} + \Xi^T\Xi)(I_{p\times p} + \Xi^T\Xi)^{-\frac{1}{2}} \quad (25)$$

where $(I_{p\times p} + \Xi^T\Xi) \in \mathbb{R}^{p\times p}$. Eigenvalue decomposition of $(I_{p\times p} + \Xi^T\Xi)$ can get:

$$(I_{p\times p} + \Xi^T\Xi) = Q\Lambda Q^{-1} \quad (26)$$

where $\Lambda$ is a diagonal matrix, $Q^T Q = I_{p\times p}$.

So, we have:

$$R_W(\Xi)^T R_W(\Xi) = Q\Lambda^{-\frac{1}{2}}Q^T Q\Lambda Q^T Q\Lambda^{-\frac{1}{2}}Q^T = Q\Lambda^{-\frac{1}{2}}\Lambda\Lambda^{-\frac{1}{2}}Q^T = QQ^T = I_{p\times p} \quad (27)$$

which satisfies the orthogonal constraints. ∎

The transformation $\mathcal{T}.(\cdot)$ in Step 6 is a vector transport, which is a smooth mapping to transport a tangent vector $H_{(k-1)}$ at previous points $M_{(k-1)}$ to a tangent vector at points $M_{(k)}$. For the tangent space of the Euclidean manifold is itself, the vector transport is an identity function $\mathcal{T}_W(\cdot) = \mathrm{id}_W$.

In Stiefel manifold $\mathrm{St}(1+n+n^2,p)$, the vector transport is:

$$\mathcal{T}_{\widetilde{W}_{(k)}}(H_{(k-1)}) = (I - \widetilde{W}_{(k)}\widetilde{W}_{(k)}^T)H_{(k-1)} \quad (28)$$

Theorem 2 proves that the vector of iteration points is on its tangent space after the vector transport.

**Theorem 2:** Let vector transport be $\mathcal{T}_{\widetilde{W}_{(k)}}(H_{(k-1)}) = (I - \widetilde{W}_{(k)}\widetilde{W}_{(k)}^T)H_{(k-1)}$, the vector after vector transport is on the tangent space, i.e., satisfies:

$$\widetilde{W}_{(k)}^T \mathcal{T}_{\widetilde{W}_{(k)}}(H_{(k-1)}) + \mathcal{T}_{\widetilde{W}_{(k)}}(H_{(k-1)})^T \widetilde{W}_{(k)} = 0 \quad (29)$$

**Proof.** For the parameter $\widetilde{W}_{(k)}$ is calculated by: $\widetilde{W}_{(k)} = R_{\widetilde{W}_{(k-1)}}\left(t^* H_{(k-1)}\right)$, we substitute (21) into (29):

$$\widetilde{W}_{(k)}^T \mathcal{J}_{\widetilde{W}_{(k)}}\left(H_{(k-1)}\right) + \mathcal{J}_{\widetilde{W}_{(k)}}\left(H_{(k-1)}\right)^T \widetilde{W}_{(k)}$$
$$= a^{-\frac{1}{2}}(b - (\widetilde{W}_{(k-1)}^T \widetilde{W}_{(k-1)} + tH_{(k-1)}^T \widetilde{W}_{(k-1)} + t\widetilde{W}_{(k-1)}^T H_{(k-1)} + t^2 H_{(k-1)}^T H_{(k-1)})a^{-1}b) + (b^T - b^T a^{-1}(\widetilde{W}_{(k-1)}^T \widetilde{W}_{(k-1)} + tH_{(k-1)}^T \widetilde{W}_{(k-1)} + t\widetilde{W}_{(k-1)}^T H_{(k-1)} + t^2 H_{(k-1)}^T H_{(k-1)}))a^{-\frac{1}{2}} \quad (30)$$

where $a = \left(I_{p\times p} + t^2 H_{(k-1)}^T H_{(k-1)}\right)$, $b = \left(\widetilde{W}_{(k-1)}^T H_{(k-1)} + tH_{(k-1)}^T H_{(k-1)}\right)$.

Since $H_{(k-1)}$ is a tangent vector at $\widetilde{W}_{(k-1)}$, we have $\widetilde{W}_{(k-1)}^T H_{(k-1)} + H_{(k-1)}{}^T \widetilde{W}_{(k-1)} = 0$. So

$$\widetilde{W}_{(k)}^T \mathcal{J}_{\widetilde{W}_{(k)}}\left(H_{(k-1)}\right) + \mathcal{J}_{\widetilde{W}_{(k)}}\left(H_{(k-1)}\right)^T \widetilde{W}_{(k)}$$
$$= a^{-\frac{1}{2}}\left(b - \left(\widetilde{W}_{(k-1)}^T \widetilde{W}_{(k-1)} + t^2 H_{(k-1)}^T H_{(k-1)}\right)a^{-1}b\right) + \left(b^T - b^T a^{-1}\left(\widetilde{W}_{(k-1)}^T \widetilde{W}_{(k-1)} + t^2 H_{(k-1)}^T H_{(k-1)}\right)\right)a^{-\frac{1}{2}} \quad (31)$$

Since $\widetilde{W}_{(k-1)}$ is a point on the Stiefel manifold, we have $\widetilde{W}_{(k-1)}^T \widetilde{W}_{(k-1)} = I_{p\times p}$, and $\left(\widetilde{W}_{(k-1)}^T \widetilde{W}_{(k-1)} + t^2 H_{(k-1)}^T H_{(k-1)}\right) = \left(I_{p\times p} + t^2 H_{(k-1)}^T H_{(k-1)}\right) = a$.

Then we have $\widetilde{W}_{(k)}^T \mathcal{J}_{\widetilde{W}_{(k)}}\left(H_{(k-1)}\right) + \mathcal{J}_{\widetilde{W}_{(k)}}\left(H_{(k-1)}\right)^T \widetilde{W}_{(k)} = a^{-\frac{1}{2}}(b - aa^{-1}b) + (b^T - b^T a^{-1} a)a^{-\frac{1}{2}} = 0$ ∎

In Step 6, $\gamma$ means the direction parameter. We confine $\gamma$ to the following formula, which is proposed in [24]:

$$\gamma = \frac{< G_{(k)}, G_{(k)} - G_{(k-1)} >}{< H_{(k-1)}, G_{(k-1)} >} \quad (32)$$

where the inner product $\langle\cdot,\cdot\rangle$ is defined as:

$$\langle (P_1, Q_1), (P_2, Q_2) \rangle = \mathrm{tr}\left({P_1}^T P_2\right) + \mathrm{tr}\left({Q_1}^T Q_2\right) \quad (33)$$

**Remark 1:** The linear conjugate gradient algorithm provides a super-linear convergence rate, but the convergence rate of the geometric conjugate gradient algorithm on the manifold is interesting and complex. Research works [23] have shown the convergence behavior of the geometric conjugate gradient algorithm, but the convergence rate is not fully discussed. We studied it experimentally. For more details, please refer to Section V-B. The convergence plot is shown in Fig. 9.

**Remark 2:** In [10], the reduced rank Procrustes rotation was used to solve the parameters with orthogonal constraints. However, an alternate optimization strategy was adopted greedily with the risk of oscillation in the optimization [26]. Moreover, due to the reduced rank Procrustes rotation, the final solution cannot be guaranteed to minimize the original optimization problem [10].

### C. *Motivation Example*

In order to analyze the performance of SCA in feature extraction, we first introduce a toy heteroscedastic process, which is given by:

$$\begin{cases} x_1 = t_1 + e_1 \\ x_2 = t_1^3 - 4.5t_2^2 + 6t_1 + t_2 + e_2 \\ x_3 = 3t_1^4 - t_2^3 + 3t_2^2 + e_3 \end{cases} \quad (34)$$

where $t_1$ and $t_2$ are independently sampled from the standard normal distribution $\mathcal{N}(0,1)$. $e_1$, $e_2$ and $e_3$ are independent variables representing the variance of $x_1$, $x_2$ and $x_3$. For the training data, $e_1$, $e_2$ and $e_3$ are sampled from the normal distribution $\mathcal{N}(0, 0.1)$, while $e_1$, $e_2$ and $e_3$ are sampled from the normal distribution $\mathcal{N}(0, 0.5)$ for the testing data. In this design, the training data and the testing data are heteroscedastic.

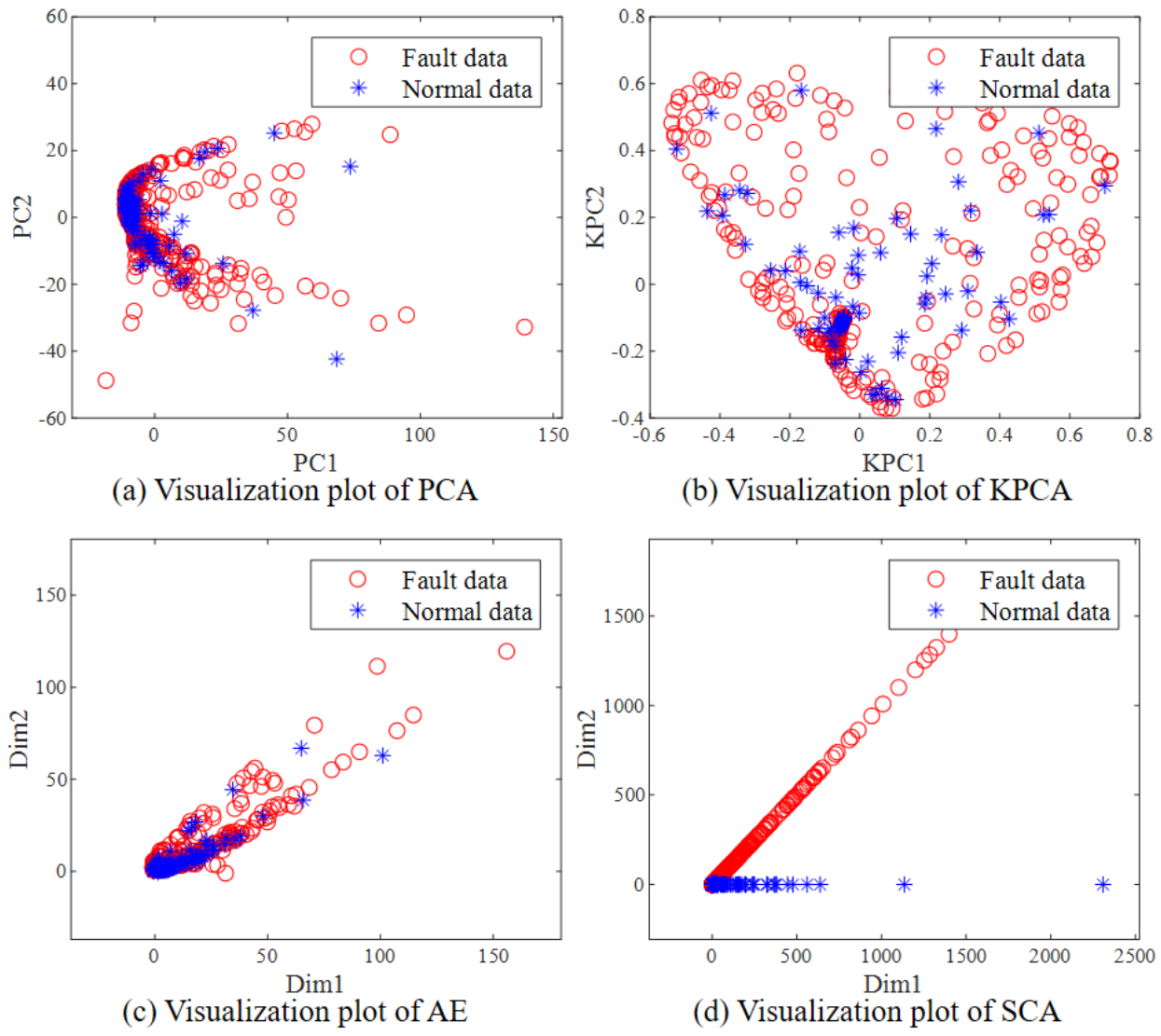

(a) Visualization plot of PCA (b) Visualization plot of KPCA (c) Visualization plot of AE (d) Visualization plot of SCA

Fig. 4. Visualization of the normal and the fault samples of the toy process on the first two dimensions of PCA, KPCA, AE, and SCA.

Training data and testing data are generated according to (34). Training data, including 500 samples, is set to be mean and unit variance. In the testing data, the first 100 samples are normal; the other 400 samples are fault by adding 1 to $x_i$ $(i = 1, 2, 3)$.

Fig. 4 shows the visualization results of the extracted features in two-dimensional spaces by PCA, KPCA, AE, and our proposed SCA, respectively, for the testing data. The blue "∗" denotes normal samples, and the red "∘" denotes fault samples. It can be found in Fig. 4 (a-c) that the normal samples and the fault samples largely overlap. However, in Fig. 4 (d), the degree of overlap between the normal sample and the faulty sample is much smaller than Fig. 4 (a-c). It means that, compared with PCA, KPCA, and AE, SCA is more suitable to handle this heteroscedastic data.

## IV. Fault Detection Based on SCA

In this section, we develop a new fault detection method based on SCA. The implementation procedure is as follows. First, in the modeling phase, process data is collected for each variable under normal conditions. Then using SCA to obtain the nonlinear feature extraction of neural network $g(\boldsymbol{x}; W)$. Finally, Hotelling $T^2$ statistic is calculated for fault detection.

Let $\boldsymbol{g}^{(i)} = g(\boldsymbol{x}^{(i)}; W)$ be a potential nonlinear feature of $\boldsymbol{x}^{(i)} (i = 1,2,\cdots,m)$. $\Sigma_g$ is a covariance matrix of features $G = \left[\boldsymbol{g}^{(1)}, \boldsymbol{g}^{(2)}, \cdots \boldsymbol{g}^{(m)}\right]$. The $T^2$ statistic of $\boldsymbol{g}^{(i)}$ is calculated as follows:

$$T^{2^{(i)}} = \boldsymbol{g}^{(i)} {\Sigma_g}^{-1} {\boldsymbol{g}^{(i)}}^T \quad (35)$$

Because $\boldsymbol{g}^{(i)}$ has no prior knowledge available, the confidence bounds of $T^2$ statistic are calculated by kernel density estimation (KDE) [25]. Let $T^{2^{(1)}}, T^{2^{(2)}}, \cdots, T^{2^{(m)}}$, whose density $\rho(\cdot)$ is unknown, be the $T^2$ statistic of $\boldsymbol{g}^{(1)}, \boldsymbol{g}^{(2)}, \cdots \boldsymbol{g}^{(m)}$. The kernel density estimation for $T^2$ statistic is:

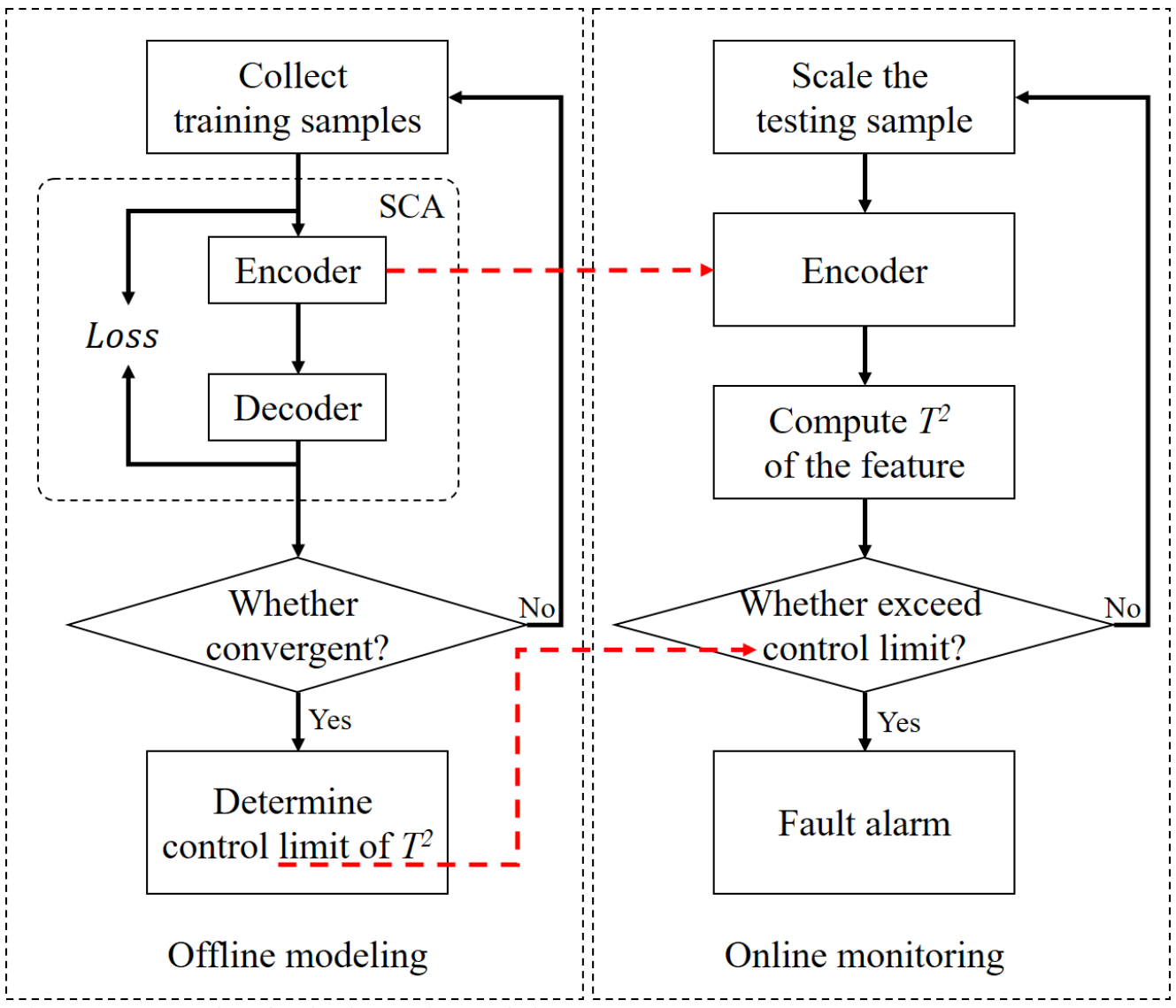

Fig. 5. Steps for fault detection of the SCA.

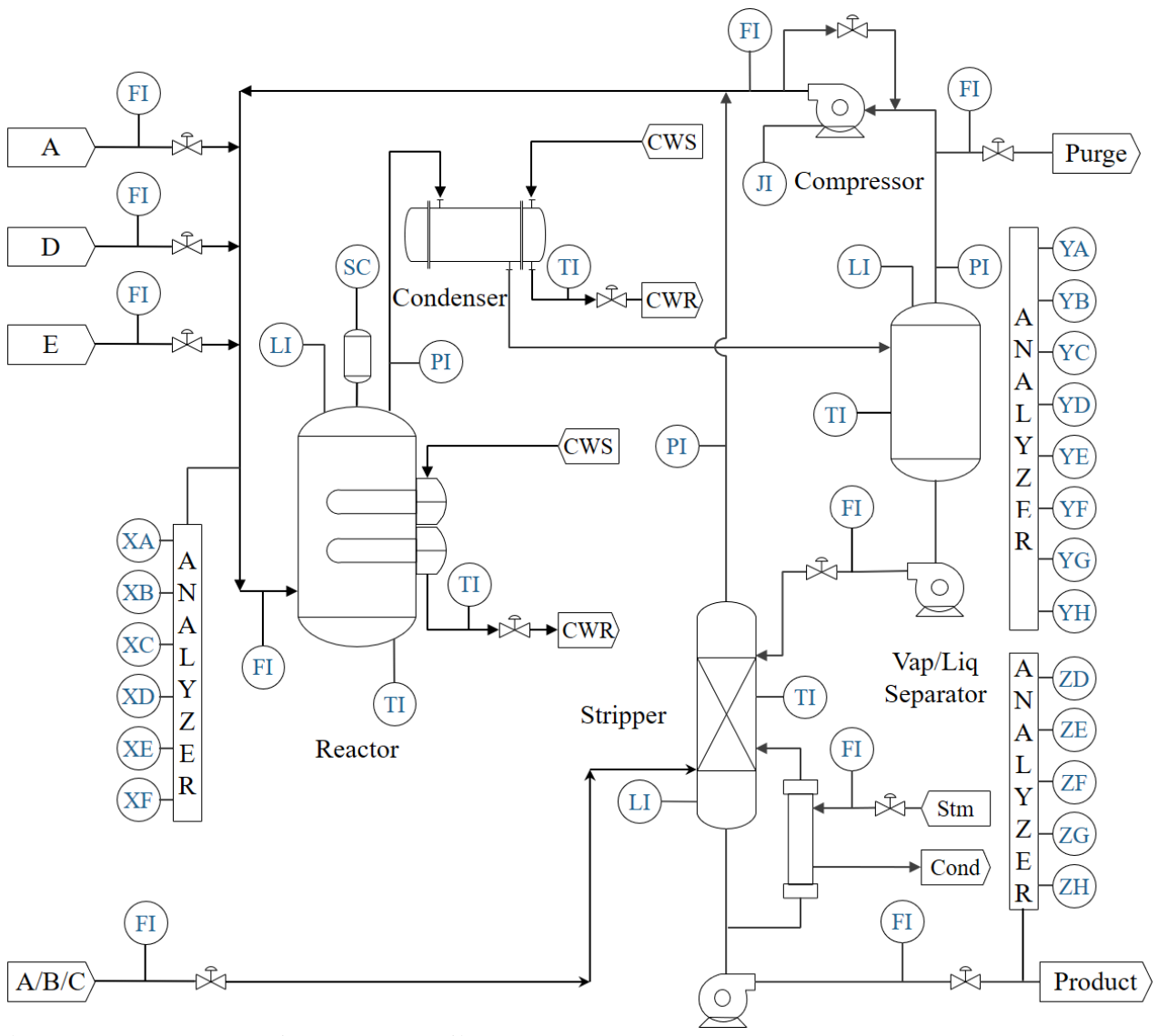

Fig. 6. TEP control structure diagram.

$$\hat{\rho}(T^2) = \frac{1}{N}\sum_{i=1}^{N} K\left(T^2 - {T_i}^2\right) = \frac{1}{hN}\sum_{i=1}^{N} K\left(\frac{T^2 - {T_i}^2}{h}\right) \quad (36)$$

where $K(\cdot)$ denotes a non-negative integrable function, $h > 0$ is the bandwidth parameter. Density estimation using the Gaussian kernel is given by:

$$\hat{\rho}(T^2) = \frac{1}{\sqrt{2\pi}hN}\sum_{i=1}^{N} \exp\left(\frac{\left(T^2 - {T_i}^2\right)^2}{2h^2}\right) \quad (37)$$

After estimating the density function $\hat{\rho}(T^2)$, integrating $\hat{\rho}_h(T^2)$ over a continuous range $[0, \tau]$ gives the probability,

$$P(T^2 \le \tau) = \int_0^{\tau} \hat{\rho}(T^2)\mathrm{d}T^2 \quad (38)$$

Let $\varsigma$ be a specified significance level (In this paper, $\varsigma = 0.01$.), the control limit of $T^2$ can be calculated from:

$$\int_0^{\tau_{T^2}} \hat{\rho}(T^2)\mathrm{d}T^2 = \varsigma \quad (39)$$

In order to test $T^{2(new)}$ statistic, it is necessary to check the following situation: if $T^{2(new)} < \tau_{T^2}$, then $\boldsymbol{x}^{(new)}$ is normal; otherwise $\boldsymbol{x}^{(new)}$ is abnormal.

The offline model and online monitoring flowchart are shown in Fig. 5. The offline model and online monitoring procedures are as follows:

**Offline model:**

(1) Collect samples from normal processes as training data.

(2) Normalize training data to zero mean and unit variance.

(3) Initialize $W$, $\widetilde{W}$.

(4) Calculate $W$ and $\widetilde{W}$ by Algorithm 1.

(5) Go to step 4 if not converge; otherwise, go to step 7.

(6) Calculate $T^2$ statistic by (37).

(7) Determine the control limits of $T^2$ by (39).

**Online monitoring:**

(1) Collect new samples $\boldsymbol{x}^{(new)}$. Normalize $\boldsymbol{x}^{(new)}$ according to the parameters of the training data.

(2) Extract features $\boldsymbol{g}^{(new)} = g(\boldsymbol{x}^{(new)}; W)$.

(3) Compute the $T^2$ statistic of $\boldsymbol{g}^{(new)}$.

(4) Alarm if the $T^2$ statistic exceeds the control limit; otherwise, treat $\boldsymbol{x}^{(new)}$ as a normal data.

TABLE I
TENNESSEE-EASTMAN PROCESS (TEP) FAILURE MODE

| Fault No. | Description | Type |
|---|---|---|
| 1 | A/C feed ratio, B composition constant (Stream 4) | Step |
| 2 | B composition, A/C ratio constant (Stream 4) | Step |
| 3 | D feed temperature (Stream 2) | Step |
| 4 | Reactor cooling water inlet temperature | Step |
| 5 | Condenser cooling water inlet temperature | Step |
| 6 | A feed loss (Stream 1) | Step |
| 7 | C header pressure loss (Stream 4) | Step |
| 8 | A, B, C feed composition (Stream 4) | Random variation |
| 9 | D feed temperature (Stream 2) | Random variation |
| 10 | C feed temperature (Stream 4) | Random variation |
| 11 | Reactor cooling water inlet temperature | Random variation |
| 12 | Condenser cooling water inlet temperature | Random variation |
| 13 | Reaction kinetics | Slow drift |
| 14 | Reactor cooling water valve | Sticking |
| 15 | Condenser cooling water valve | Sticking |
| 16 | Unknown | Unknown |
| 17 | Unknown | Unknown |
| 18 | Unknown | Unknown |
| 19 | Unknown | Unknown |
| 20 | Unknown | Unknown |
| 21 | Valve (Stream 4) | Constant position |

## V. SIMULATION AND DISCUSSION

Tennessee-Eastman Process (TEP), founded by Eastman Chemical Company, provides a realistic industrial process for evaluating process control and monitoring methods [27]. Based on real industrial processes simulation, TEP contains 30 equations and 148 algebraic equations, generating nonlinear, strongly coupled, and dynamic data. TEP has five main units: reactor, condenser, compressor, separator, and stripper; its control structure is shown in Fig. 6.

In the experiment, 52 variables are selected as monitoring variables, including 22 continuous process variables, 19 combinations, and 11 manipulation variables. There are 21 fault modes simulated by TEP, which are shown in Table I. The training data contains 500 normal samples. The testing data contains 960 samples; among them, the first 160 samples are normal samples, and the 161st to the 960th samples are fault samples.

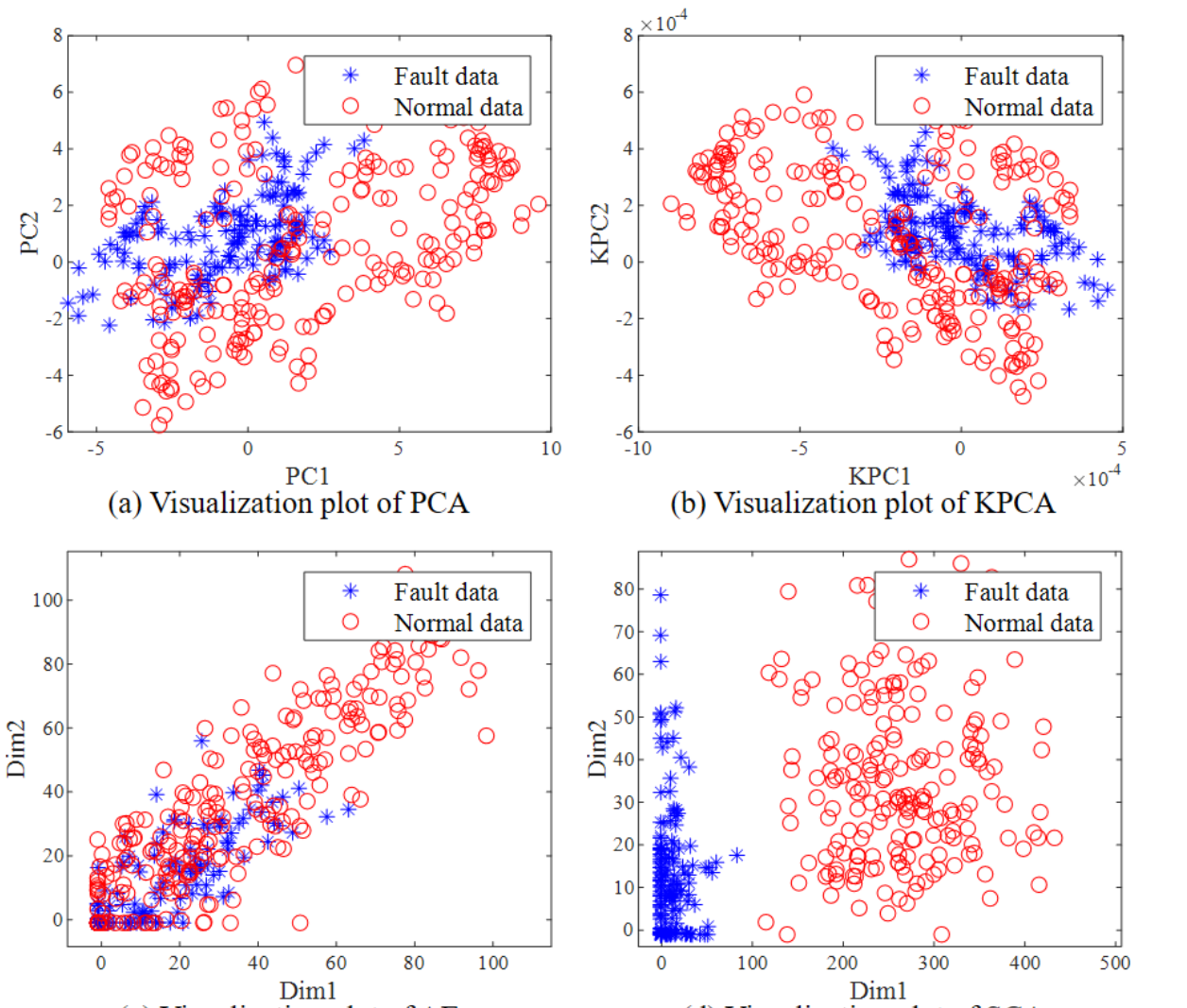

(a) Visualization plot of PCA (b) Visualization plot of KPCA

(c) Visualization plot of AE (d) Visualization plot of SCA

Fig. 7. Visualization of the normal and the fault samples of fault 4 on the first two dimensions of PCA, KPCA, AE, and SCA.

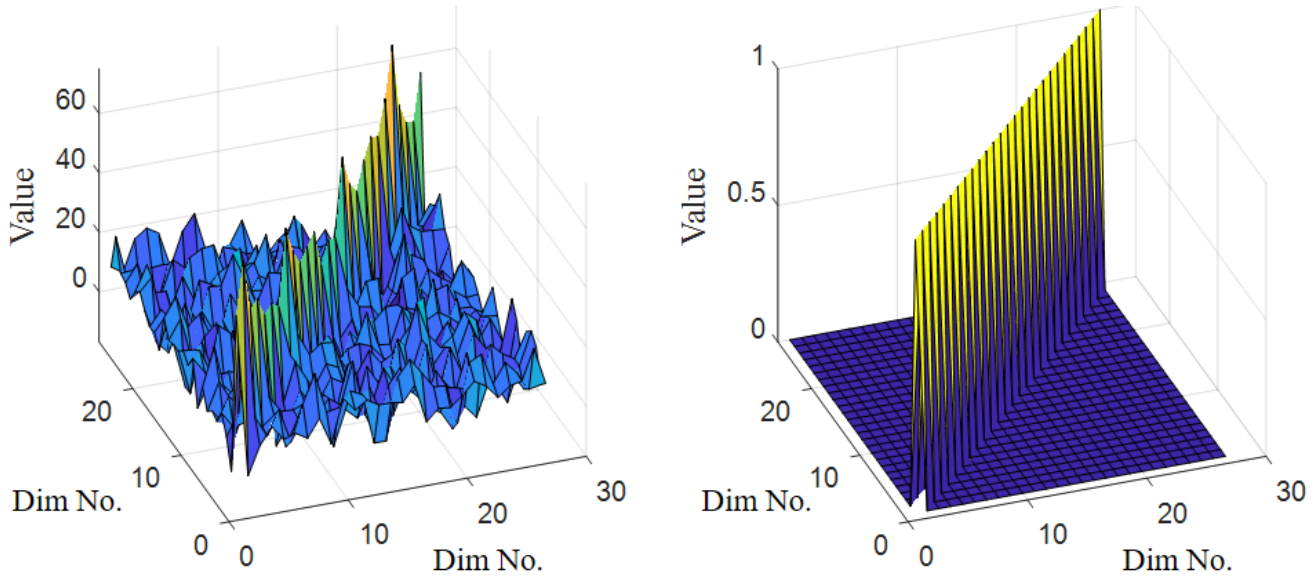

(a) AE (b) SCA

Fig. 8. Visualization of $\widetilde{W}^T\widetilde{W}$ of the AE and SCA.

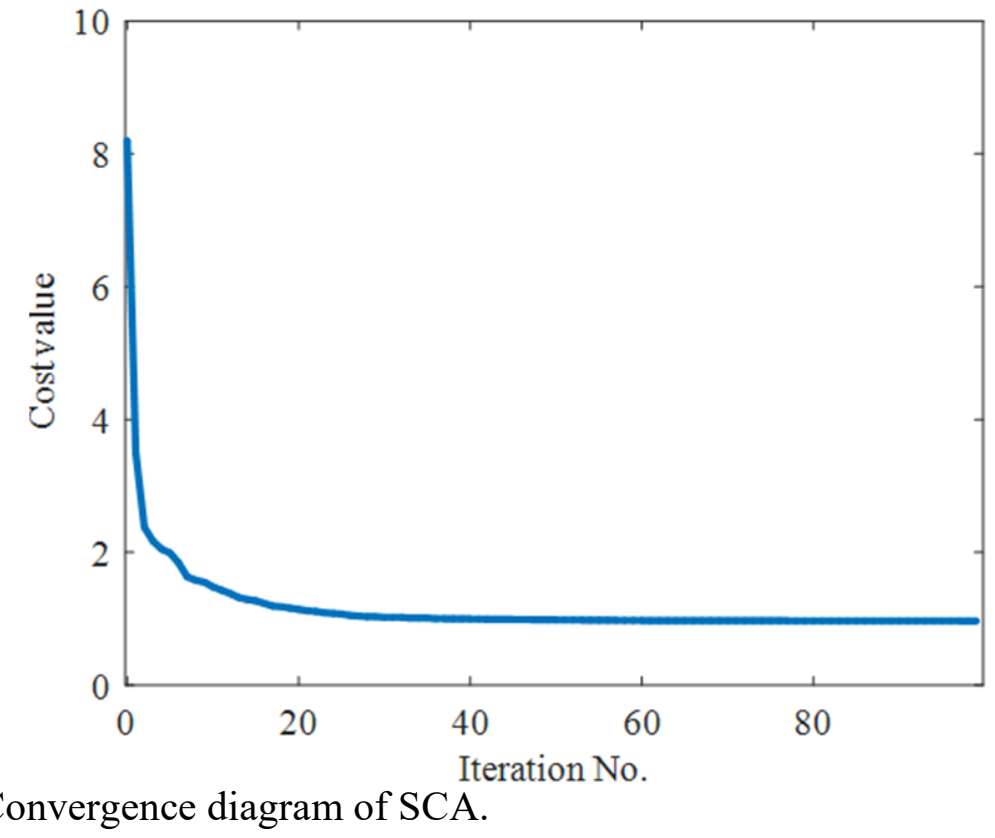

Fig. 9. Convergence diagram of SCA.

In this paper, PCA, KPCA, AE, and our proposed SCA are used for fault detection on TEP. The kernel of KPCA is Gaussian kernel $k(\boldsymbol{x}_i, \boldsymbol{x}_j) = \exp(-\|\boldsymbol{x}_i - \boldsymbol{x}_j\|^2/c)$ The kernel parameter $c$ is $10n\bar{\delta}$, where $\bar{\delta}$ denotes the average of the standard deviations of different variables.

TABLE II

MISS DETECTION RATE (MDR) AND FALSE ALARM RATE(FAR) OF PCA, KPCA, AUTOENCODER(AE), AND SCA

| Fault | PCA | | KPCA | | AE | | SCA | |
|---|---|---|---|---|---|---|---|---|
| No. | MDR | FAR | MDR | FAR | MDR | FAR | MDR | FAR |
| 1 | 0.50 | 3.75 | 0.50 | 1.88 | **0.13** | 0.63 | 0.75 | 0.00 |
| 2 | 1.75 | 1.88 | 1.75 | 3.13 | **1.25** | 1.88 | 1.75 | 0.00 |
| 3 | 96.63 | 0.00 | 97.13 | 0.63 | 96.75 | 2.50 | 97.13 | 0.00 |
| 4 | 50.38 | 0.63 | 58.50 | 1.88 | 75.38 | 1.88 | **46.63** | 0.00 |
| 5 | 72.38 | 0.63 | 74.13 | 1.88 | 73.50 | 1.88 | 74.13 | 0.00 |
| 6 | 1.00 | 0.63 | 0.88 | 1.25 | **0.00** | 0.00 | **0.00** | 0.63 |
| 7 | **0.00** | 0.00 | **0.00** | 0.63 | 3.00 | 1.88 | **0.00** | 0.00 |
| 8 | **2.63** | 1.88 | **2.63** | 0.63 | 2.88 | 1.25 | **2.63** | 0.00 |
| 9 | 96.25 | 3.75 | 97.88 | 1.88 | 97.38 | 4.38 | 96.75 | 5.63 |
| 10 | 63.88 | 0.63 | 69.88 | 1.25 | 62.63 | 0.00 | 56.18 | 0.00 |
| 11 | **45.38** | 1.88 | 50.13 | 1.25 | 49.88 | 0.63 | 46.38 | 0.63 |
| 12 | **1.00** | 2.50 | **1.00** | 1.25 | 4.50 | 1.25 | 1.63 | 0.00 |
| 13 | 4.88 | 0.63 | **4.50** | 2.50 | 5.88 | 0.63 | 5.25 | 0.00 |
| 14 | **0.13** | 1.25 | 0.38 | 1.25 | 0.75 | 1.88 | **0.13** | 0.00 |
| 15 | 94.25 | 1.88 | 94.88 | 0.63 | 96.38 | 0.63 | 94.38 | 0.00 |
| 16 | 80.13 | 5.00 | 83.75 | 4.38 | 70.88 | 4.38 | 70.88 | 9.38 |
| 17 | 19.25 | 1.88 | 21.63 | 1.25 | 17.38 | 1.88 | **16.88** | 0.00 |
| 18 | 10.88 | 1.88 | 11.13 | 1.25 | 11.00 | 1.88 | **10.75** | 0.00 |
| 19 | 89.00 | 1.25 | 91.75 | 2.50 | 93.00 | 0.63 | 98.00 | 0.63 |
| 20 | 63.75 | 0.63 | 66.00 | 1.25 | 63.25 | 0.00 | 55.75 | 0.00 |
| 21 | 61.00 | 0.63 | 60.63 | 1.25 | 68.88 | 1.25 | 63.75 | 1.25 |
| Best | 5 | | 4 | | 4 | | 7 | |

### *A. Visualization of Different Methods*

To represent the features extracted by different methods intuitively, we use each method to extract two features of the samples of Fault 4 and plot them in Fig. 7, in which the blue "*" denotes normal samples and the red "∘" denotes fault samples. As shown in Fig. 7 (a-c), normal samples and fault samples overlap largely. However, in Fig. 7 (d), the overlap between normal samples and fault samples is much smaller than in Fig. 7 (a-c). It shows that SCA can extract key features for detecting Fault 4.

Fig. 8 shows the difference of $\widetilde{W}$ between AE and SCA. Fig. 8 (a) shows $\widetilde{W}^T\widetilde{W}$ of AE, which can be found that $\widetilde{W}$ of AE has no orthogonal constraints. Fig. 8 (b) shows $\widetilde{W}^T\widetilde{W}$ of SCA. It is easy to find $\widetilde{W}$ of SCA contains orthogonal constraints, i.e., $W^TW = I_{p\times p}$.

We experimented on a computer equipped with Intel® Core™ i5-9600KF CPU, NVIDIA® GeForce RTX™ 2060 GPU, and 16G memory. Fig. 9 plots the cost value during the training phase. The total training iterations for SCA is 228 iterations; the total training time is 4.96 seconds. For comparison, the training iterations and time of AE are 56 iterations and 0.59 seconds. Nevertheless, given the excellent performance of SCA, this time consuming is acceptable.

### *B. Case Studies*

In this section, the performance of the proposed SCA is compared with PCA, KPCA, and AE. In order to retain 85% of the energy in the eigenspectrum (computed as the sum of eigenvalues), we set the PCA dimension of features $p$ to 27. For fair comparison, $p$ of KPCA, SCA, and AE use the same value for feature extraction. The missed detection rate (MDR) refers to the ratio of abnormal events misidentified as normal events during monitoring. The false alarm rate (FAR) refers to the ratio of normal events misidentified as abnormal events. To detect all

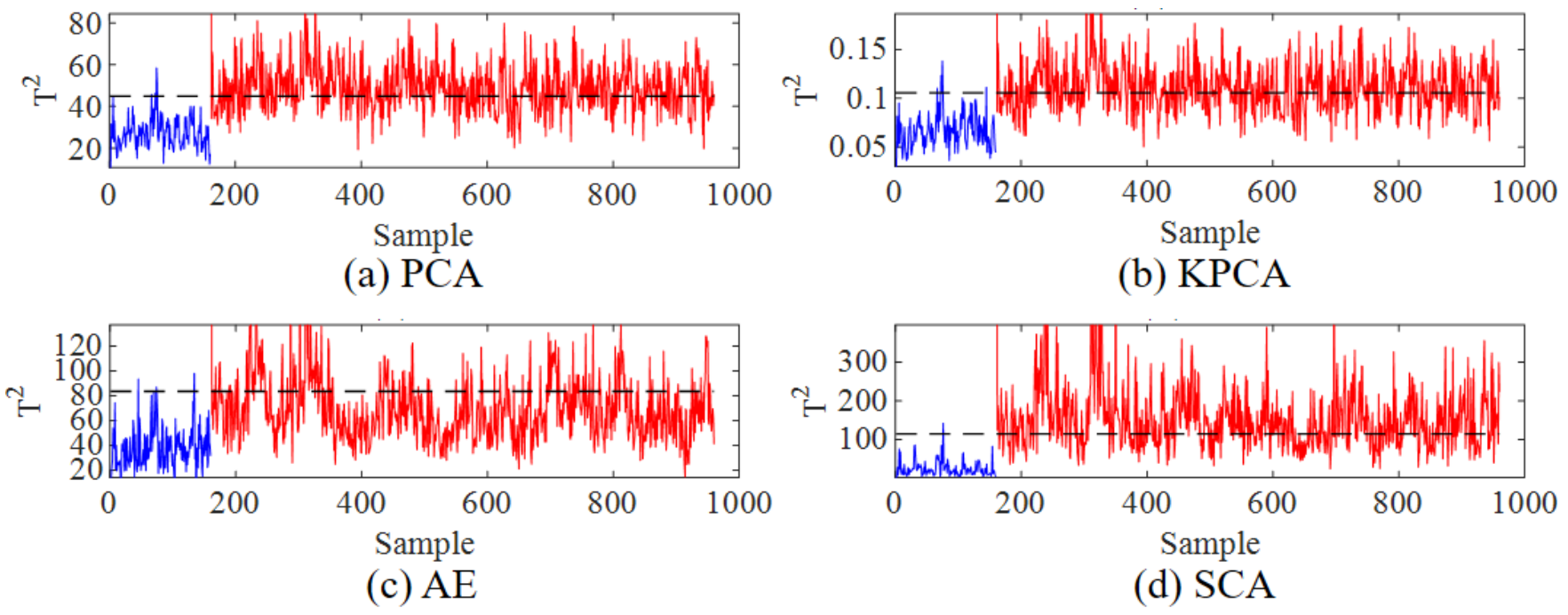

Fig. 10. Results of Fault 4 monitoring using four different methods.

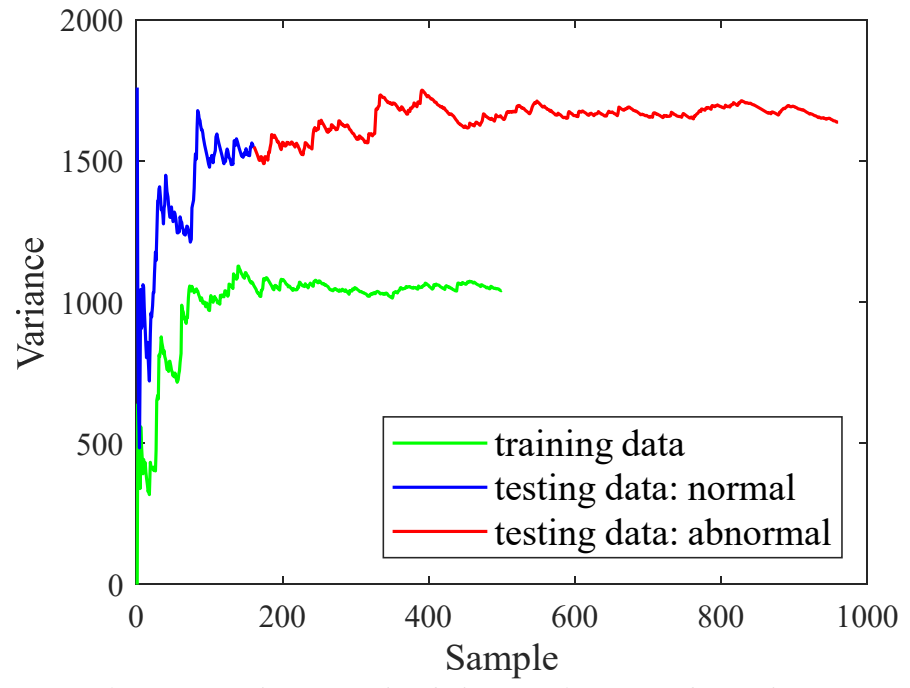

Fig. 11. The trend of the variance of Fault 4.

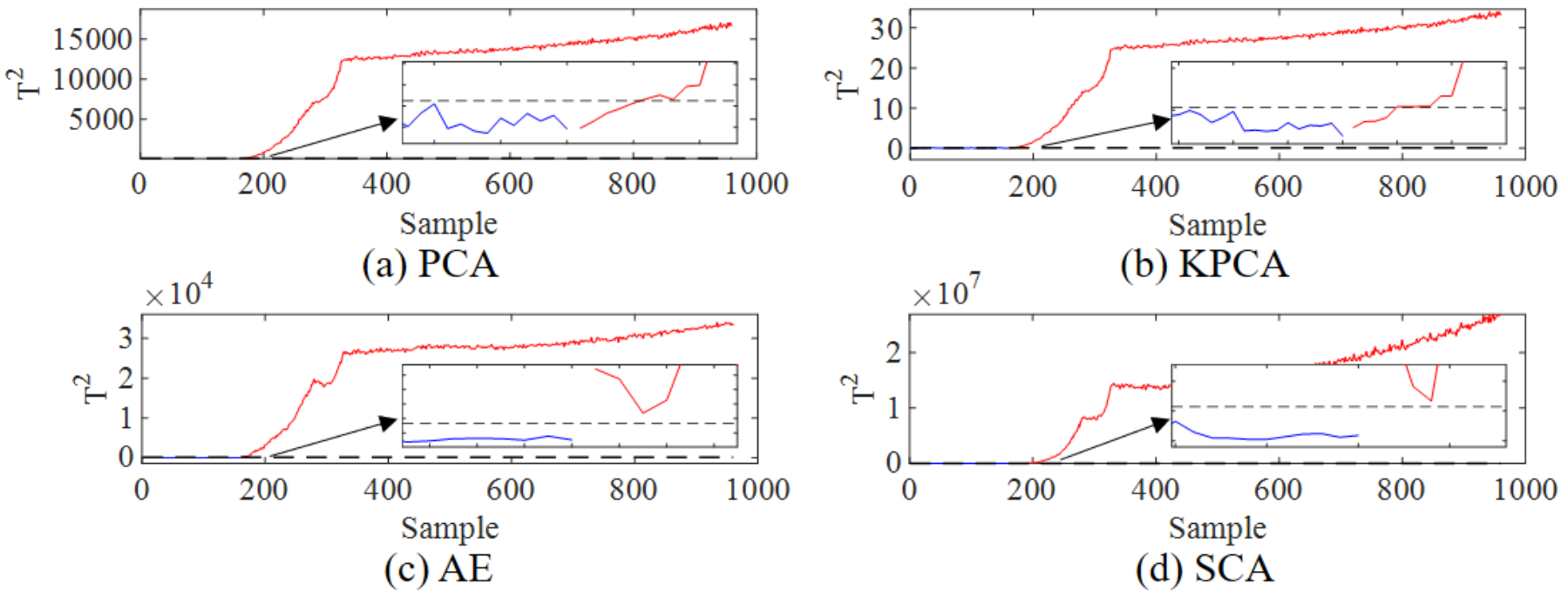

Fig. 12. Results of Fault 6 monitoring using four different methods.

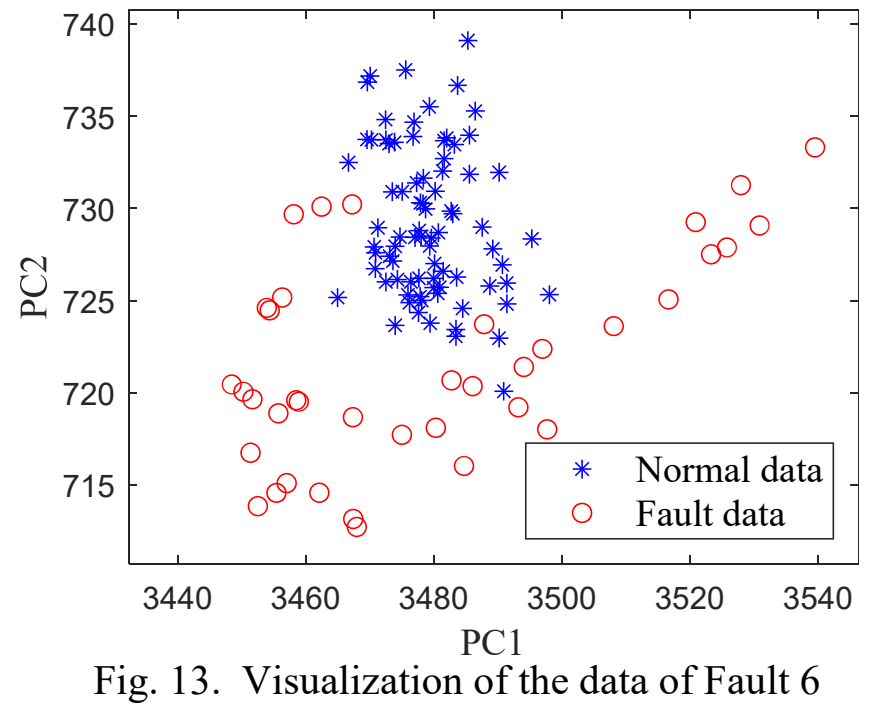

Fig. 13. Visualization of the data of Fault 6

21 faults, MDR and FAR are recorded together in Table II. MDR or FAR with smaller values indicates better performance. In Table II, the best performance for each fault is highlighted in bold. In this study, only fault conditions with MDR $< 50\%$ and FAR $\leq 5\%$ is considered. Because if MDR is larger than 50%, the detection performance is even worse than the random selection whose MDR is 50%. The threshold of FAR commonly used in fault detection is 5% [27].

SCA obtains the lowest case of MDR in 7 cases. PCA gives 5 best cases, KPCA gives 4 best cases, AE gives 4 cases.

For no significant change in mean-variance and other statistical data by comparing the observed variables related to Fault 3, 9, and 15 with those related to normal operation, it is difficult to detect these faults using multivariate statistical fault detection methods [27]. For PCA, KPCA, AE, and SCA are all unsupervised learning methods that only depend on normal training samples and their main statistical characteristics. They cannot correctly detect these faults. According to Table II, KPCA, AE, and SCA have better or the same performance as PCA in all detectable fault modes. KPCA, AE, and SCA are all nonlinear methods, which have better performance than PCA because they can better represent the original data. This means that effective subspace, residual subspace, and related statistical data are more suitable for fault detection through nonlinear transformation. Although both AE and SCA are nonlinear methods, SCA performs better than AE because AE does not contain orthogonal constraints and tends to overfit. Besides useful information, the coded features also contain a lot of noise or outliers. Although both KPCA and SCA contain orthogonal constraints in feature extraction, the performance of SCA is much better than that of KPCA. The reason is that SCA can effectively solve the case that input variables are heteroscedastic.

Fig. 10, Fig. 12, and Fig. 14 detail the fault detection results for Fault 4, 6, and 7. In these figures, the blue dots represent the first 160 normal samples; the red dots represent the remaining 800 fault samples; the black dotted line is control restrictions based on the threshold $\tau$. Blue dots above the control limit cause false alarm, while red dots below the control limits lead to missed detection.

Fig. 10 shows the results of Fault 4 by four methods. The experimental results of Fault 4 demonstrate that SCA has a good experimental effect on heteroscedastic data. The MDR of the $T^2$ statistic of PCA, KPCA, AE, and SCA is 50.38%, 58.50%, 75.38%, and 46.63%, respectively. Among them, the MDR of SCA is much smaller than the other three methods. This is because the input of SCA has second-order terms, making SCA far better at processing heteroscedastic data than the other three methods.

In order to represent the heteroscedasticity of Fault 4 intuitively, we use PCA to reduce the dimension of Fault 4 and calculate the variance of the first principal component, as shown in Fig. 11. From Fig. 11, it is easy to find that the variance between the training data and the testing data, the normal samples and the fault samples are different. According to Section II-C, high-order neural networks have natural advantages in dealing with heteroscedastic data. Features extracted by PCA, KPCA, and AE cannot deal with heteroscedastic data, but SCA has the second-order term, dealing with heteroscedastic data. The second-order term makes SCA better than the other three methods in processing such data.

Fig. 12 illustrates the results of Fault 6. The experimental results of Fault 6 show that all three nonlinear methods (KPCA, AE, and SCA) have good performances on this fault mode. The MDRs of the $T^2$ statistic of PCA, KPCA, AE, and SCA are 1.00%, 0.88%, 0.00%, and 0.00%, respectively. The results of KPCA, AE, and SCA are all better than PCA. This is because they are nonlinear methods and suitable for nonlinear data process data.

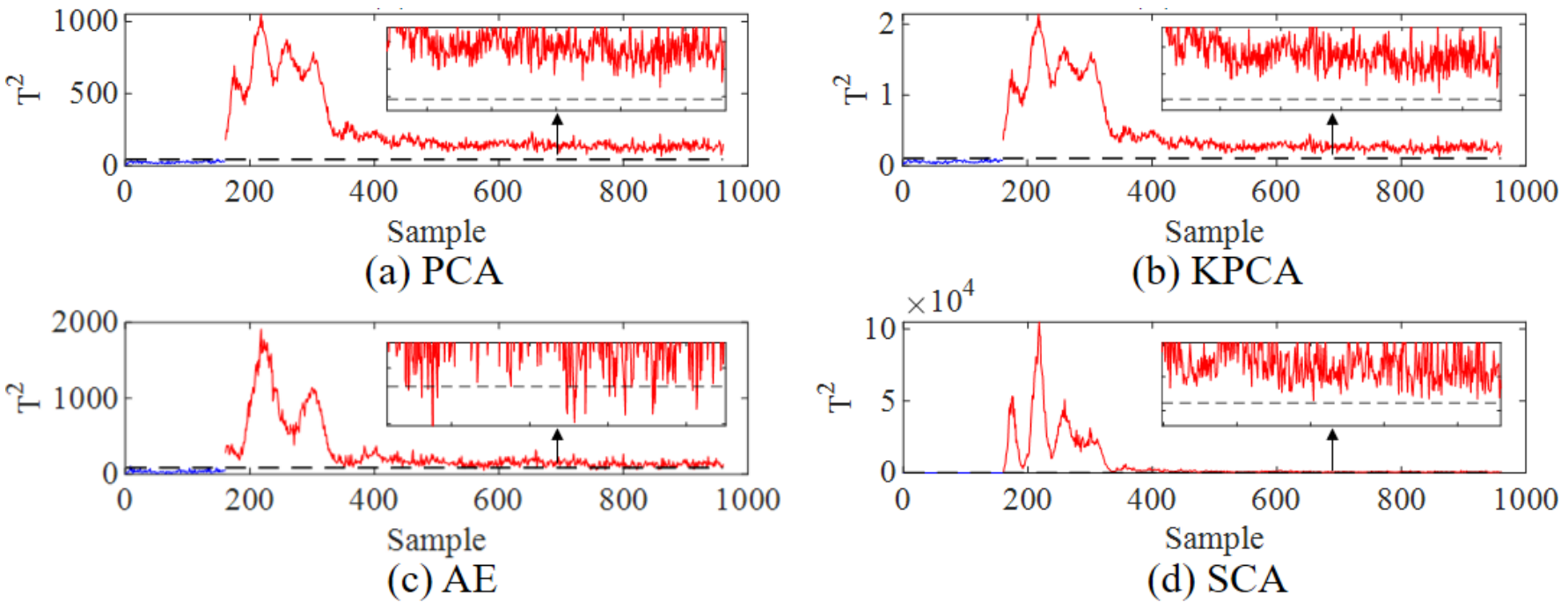

Fig. 14. Results of Fault 7 monitoring using four different methods.

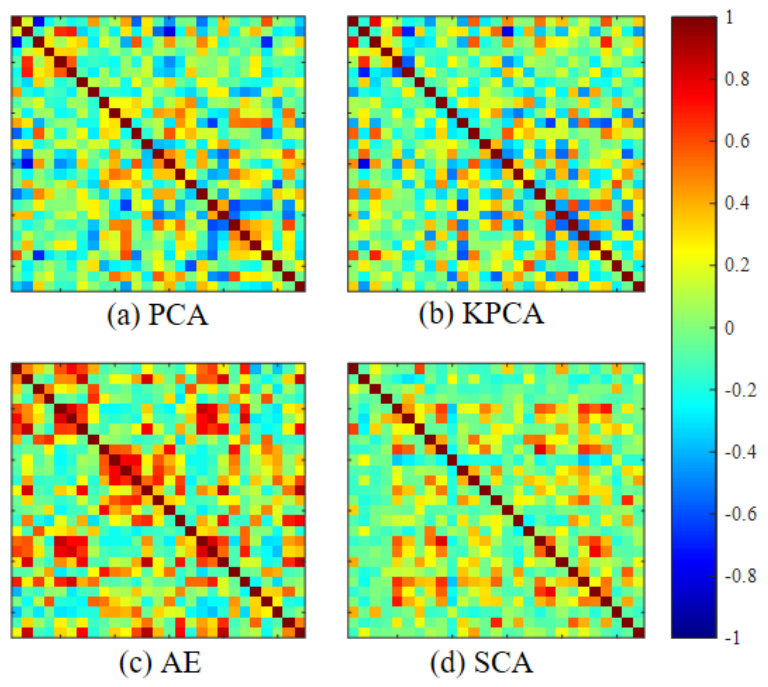

Fig. 15. The covariance matrix of the features

In order to demonstrate the nonlinearity of Fault 6, we reduced the dimension of the data of Fault 6 for visualization which can be found in Fig. 13. It can be seen that the normal samples and the fault samples are interleaved with each other and cannot be effectively detected by the linear fault detection method (PCA). However, nonlinear methods are much suitable for this fault mode. The MDR of KPCA is 0.88%, which is better than PCA, thanks to the kernel techniques to make it a nonlinear method. Both AE and SCA have the best performance (0.00%). Because they are neural network methods, they have more parameters than kernel methods (KPCA), and they are suitable for complex nonlinear processes.

Fig. 14 shows the results of the four methods of Fault 7. The experimental results of Fault 7 show that orthogonal constraints can play an essential role in this fault. the MDR of the $T^2$ statistic for AE is 3.00%, which is the worst in 4 methods. The high MDR of AE should because AE does not have orthogonal constraints, and it is easy to overfit the training data. Due to the orthogonal constraints, PCA, KPCA, and SCA, the features extracted by PCA (KPCA or SCA) have low correlation and can successfully detect this fault. Fig. 15 shows the correlation coefficients of the features extracted by the four different methods. The correlation coefficients show the correlations between the different variables. In Fig. 15, the color bar on the right displays the mapping of data values into the colormap, the warmer (colder) color indicates the stronger the positive (negative) correlation. It can be found that the features extracted by AE have a high correlation.

### C. *Ablation Study*

In order to study the influence of the second-order terms and orthogonal constraints on the fault detection task, we conducted ablation experiments. We use AE as the baseline and added second-order terms to obtain the second-order autoencoder (SAE). Then we compare them with SCA. The experimental results are shown in Table III. The up arrow in the table indicates that MDR or FAR is higher than baseline (AE), indicating that the performance of the fault detection task has worsened. The down arrow indicates that MDR or FAR is lower than baseline (AE), indicating that the performance of the fault detection task has improved.

SAE has worse performances than baseline (AE) in all fault modes which can be detected. SAE has a higher MDR than AE in 10 cases; SAE has the same MDR as AE in 1 case, but in this case (Fault 6), the FAR of SAE (0.63%) is higher than that of AE (0.00%). The reason for the performance degradation of the SAE is that the second-order terms make the method complex. The parameter amount of the SAE has been greatly increased and SAE tends to overfit. In this case, noises or anomalies in the original data may also be represented in the extracted features.

TABLE III
COMPARISON OF PRUNE EXPERIMENTS. INCLUDING AUTOENCODER(AE), SECOND-ORDER-BASED AUTOENCODER(SAE), AND SCA

| Fault | AE | | SAE | | SCA | |
|---|---|---|---|---|---|---|
| No. | MDR | FAR | MDR | FAR | MDR | FAR |
| 1 | 0.13 | 0.63 | 57.88(↑) | 1.88 | 0.75(↑) | 0.00 |
| 2 | 1.25 | 1.88 | 1.63(↑) | 0.63 | 1.75(↑) | 0.00 |
| 3 | 96.75 | 2.50 | 98.25 | 0.00 | 97.13 | 0.00 |
| 4 | 75.38 | 1.88 | 71.63 | 3.75 | 46.63(↓) | 0.00 |
| 5 | 73.50 | 1.88 | 97.50 | 3.75 | 74.13 | 0.00 |
| 6 | 0.00 | 0.00 | 0.00 | 0.63 | 0.00 | 0.63 |
| 7 | 3.00 | 1.88 | 37.50(↑) | 1.88 | 0.00(↓) | 0.00 |
| 8 | 2.88 | 1.25 | 77.88(↑) | 0.63 | 2.63(↓) | 0.00 |
| 9 | 97.38 | 4.38 | 98.50 | 1.25 | 96.75 | 5.63 |
| 10 | 62.63 | 0.00 | 96.63 | 1.88 | 56.18 | 0.00 |
| 11 | 49.88 | 0.63 | 65.88(↑) | 1.25 | 46.38(↓) | 0.63 |
| 12 | 4.50 | 1.25 | 89.25(↑) | 1.88 | 1.63(↓) | 0.00 |
| 13 | 5.88 | 0.63 | 84.25(↑) | 1.88 | 5.25(↓) | 0.00 |
| 14 | 0.75 | 1.88 | 3.00(↑) | 1.25 | 0.13(↓) | 0.00 |
| 15 | 96.38 | 0.63 | 98.00 | 1.25 | 94.38 | 0.00 |
| 16 | 70.88 | 4.38 | 96.00 | 0.63 | 70.88 | 9.38 |
| 17 | 17.38 | 1.88 | 26.50(↑) | 1.88 | 16.88(↓) | 0.00 |
| 18 | 11.00 | 1.88 | 33.88(↑) | 2.50 | 10.75(↓) | 0.00 |
| 19 | 93.00 | 0.63 | 93.00 | 1.25 | 98.00 | 0.63 |
| 20 | 63.25 | 0.00 | 78.25 | 0.63 | 55.75 | 0.00 |
| 21 | 68.88 | 1.25 | 90.63 | 0.63 | 63.75 | 1.25 |

Compared with the baseline (AE), SCA has a lower MDR in 9 cases and has a higher MDR in 2 cases. It can be found that the simultaneous use of the second-order terms and orthogonal constraints can boost the performances. Combining the second-order autoencoder with orthogonal constraints can exert the powerful expressive ability and avoid network overfitting.

Based on all the experiments performed in this section, we conclude that:

1. Although it cannot provide the best performance for all different faults, SCA is superior to PCA and KPCA and AE in the number of best performances.

2. Due to the utilization of second-order terms, SCA has the best performance in handling the heteroscedastic problem.

3. With orthogonal constraints, SCA can avoid the overfitting problem.

4. For complex industrial process data, SCA is more suitable for fault detection than PCA, KPCA, and AE. Although SCA requires more training time, it is worth balancing training time and performance due to its superior performance.

## VI. Conclusion

In this paper, we propose a new fault detection method called second-order component analysis (SCA). SCA uses the structure of autoencoder and adds second-order terms to improve nonlinear mapping capabilities. Second-order terms can also solve the heteroscedastic problem of processing data. Besides, SCA adopts orthogonal constraints to reduce the overfitting problem. To solve the constrained optimization problem of SCA, we adopt a geometric conjugate gradient algorithm on the product manifold. Experimental results on TEP show that SCA can achieve much better performance than PCA, KPCA, and autoencoder. SCA can be viewed as an alternative to prevalent data-driven fault detection methods.